\def\year{2021}\relax
\documentclass[letterpaper]{article} 
\usepackage{aaai21}  
\usepackage{times}  
\usepackage{helvet} 
\usepackage{courier}  
\usepackage[hyphens]{url}  
\usepackage{graphicx} 
\usepackage{natbib}  
\usepackage{caption} 
\usepackage[switch]{lineno}
\usepackage{makecell}
\usepackage{amsmath}
\usepackage{amsfonts}
\usepackage{color}
\usepackage{multirow}
\usepackage{arydshln}
\usepackage{subcaption}

\DeclareMathOperator\arctanh{arctanh}

\usepackage[ruled,linesnumbered]{algorithm2e}
\SetKwComment{Comment}{$\triangleright$\ }{}

\title{Doubly Residual Neural Decoder:

Towards Low-Complexity High-Performance Channel Decoding}
\author{
Siyu Liao,
Chunhua Deng, 
Miao Yin, 
Bo Yuan \\
}
\affiliations{
    Department of Electrical and Computer Engineering, Rutgers University \\
    siyu.liao@rutgers.edu, chunhua.deng@rutgers.edu, miao.yin@rutgers.edu, bo.yuan@soe.rutgers.edu
}

\begin{document}

\maketitle

\begin{abstract}
Recently deep neural networks have been successfully applied in channel coding to improve the decoding performance.
However, the state-of-the-art neural channel decoders cannot achieve high decoding performance and low complexity simultaneously. To overcome this challenge, in this paper we propose doubly residual neural (DRN) decoder. By integrating both the residual input and residual learning to the design of neural channel decoder, DRN enables significant decoding performance improvement while maintaining low complexity. Extensive experiment results show that on different types of channel codes, our DRN decoder consistently outperform the state-of-the-art decoders in terms of decoding performance, model sizes and computational cost.
\end{abstract}

\section{Introduction}
Starting from Claude Shannon’s 1948 seminal paper \cite{shannon1948mathematical}, \textit{channel codes}, also known as error correction codes, have provided
data reliability for communication and storage systems in the last seven decades. Historically, every ten
years or so information theorists discovered a new channel code that approaches the ultimate channel capacity closer
than the prior ones, thereby reshaping the way that we transmit and store data. 
For instance, low-density parity check (LDPC) codes \cite{gallager1962low,mackay1996near} that was re-discovered in 1996 and polar codes \cite{arikan2009channel} that was invented in 2009 have become the adopted channel codes solution in 5G standard. Nowadays, channel codes have served as the key enablers for the dramatic advances of modern high-quality data transmission
and high-density storage systems, including but not limited to 5G air interface, deep space communication, solid-state disk (SSD), high-speed Ethernet etc.

\begin{figure}[tb]
\centering
\includegraphics[clip, trim=9cm 7cm 9.5cm 7cm, width=\columnwidth]{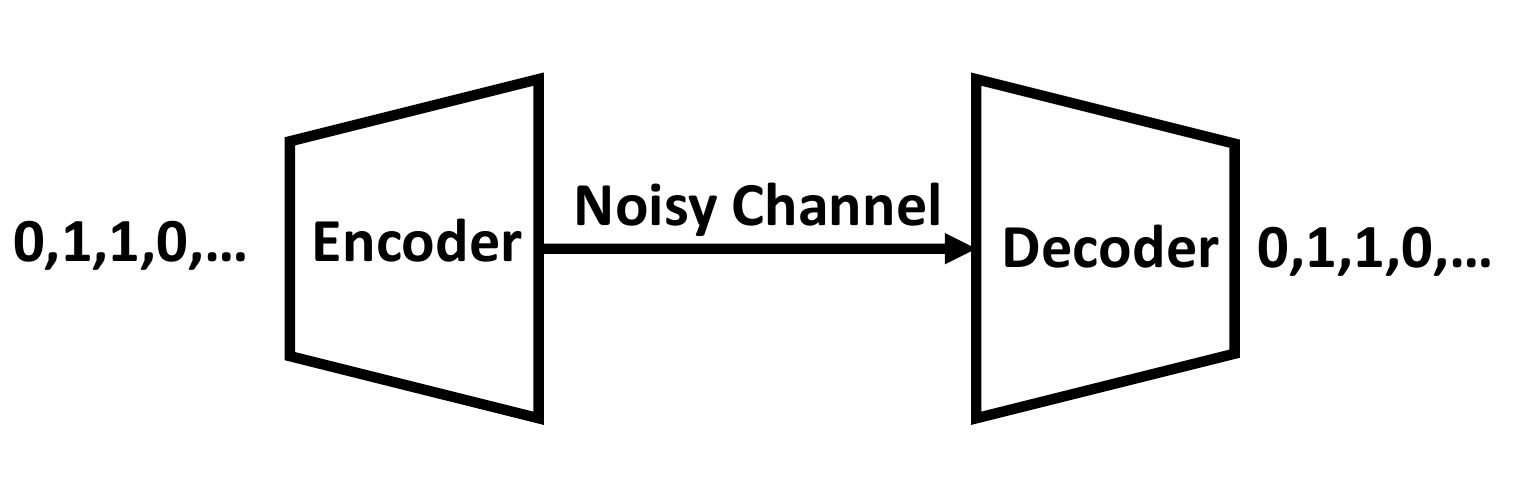}
\caption{A channel codec uses one encoder and one decoder to recover the information after noisy transmission.}
\label{fig:encdec}
\end{figure}

\textbf{Channel Encoding \& Decoding.} In general, the key idea of channel coding is to first \textit{encode} certain redundancy into the bit-level message that will be transmitted over noisy channel, and then at the receiver end to \textit{decode} the corrupted message for recovery via utilizing the redundancy information. Based on such underlying mechanism, a channel codec consists of one encoder and one decoder at the transmitter end and receiver end, respectively (see Figure \ref{fig:encdec}). In most cases, \textit{channel decoder is much more expensive than encoder in terms of both space and computational complexity}. This is because in encoding phase only simple exclusive OR operations are needed at the bit level; while in decoding phase the more advanced but complicated algorithms are needed to correct the errors occurred by noisy transmission. To date, the most popular and powerful channel decoding algorithm is \textit{iterative belief propagation (BP)} \cite{fossorier1999reduced}.

\textbf{Deep Learning for Channel Decoder.} From the perspective of machine learning, the role of channel decoder can be interpreted as a special multi-label binary classifier or denoiser. 
Based on such observation and motivated by the current unprecedented success of deep neural network (DNN) in various science and engineering applications, recently both information theory and machine learning communities are beginning to study the potential integration of deep neural network into channel codec, especially for the high-performance channel decoder design. 
A simple and natural idea along this direction is to use the classical deep autoencoder to serve as the entire channel codec \cite{o2017introduction}. 
Although this domain knowledge-free strategy can work for very short channel codes (e.g. less-than-10 code length), it cannot provide satisfied decoding performance for moderate and long channel codes, which are much more important and popular in the practical industrial standard and commercial systems.

\textbf{The State of the Art: NBP \& HGN Decoders.} 
Recently, several studies \cite{nachmani2016learning,cammerer2017scaling,gruber2017deep,lugosch2017neural,nachmani2018deep} have shown that, by integrating the existing mathematical structure and characteristics of classical decoding approach, e.g. iterative BP, these \textit{domain knowledge-based} neural channel decoders can provide promising decoding performance for the longer channel codes. 
Among those recent progress, both of the two state-of-the-art works, namely \textit{Neural BP (NBP)} decoder \cite{nachmani2018deep}  and \textit{Hyper Graph Neural (HGN)} decoder \cite{nachmani2019hyper}, are based on the "deep unfolding" methodology \cite{hershey2014deep}. 
Specifically, NBP decoder unfolds the original iterative BP decoder to the neural network format, and then trains the scaling factors instead of empirically setting. Following the similar strategy, HGN decoder further replaces the original message updating step of BP algorithm with a graph neural network (GNN) to form a hyper graph neural network (HGNN). As reported in their experiments on different types of channel codes, such proper utilization of the domain knowledge directly makes the neural channel decoders outperform the traditional BP decoder.\footnote{Some recent studies also propose to use neural networks to design new channel codes \cite{kim2018deepcode,ebada2019deep,jiang2019turbo,burth2020joint,kim2020physical}. In this paper we focus on designing neural channel decoders for the existing widely used channel codes (such as LDPC, Polar and BCH codes).}

\textbf{Limitations of Existing Works.} Despite the current encouraging progress, the state-of-the-art neural channel decoders are still facing several challenging limitations. Specifically, NBP decoder and its variants do not provide significant improvement on decoding performance over the traditional method. For some codes (e.g. Polar codes) with moderate or high code rates, the bit error rate (BER) performance improvement brought by NBP decoder is very slight. On the other hand, though HGN decoder indeed provides significant decoding gain over the conventional BP decoder -- HGN decoder currently maintains the best decoding performance among all the neural channel decoders, the hyper graph neural network structure makes the entire decoder suffer very large model size, thereby causing high storage cost and computational cost for both training and inference phases. Considering channel codes are widely used in the latency-restrictive resource-restrictive scenarios, such as mobile devices and terminals, the expensive deployment cost of HGN decoder makes it infeasible for practical applications.

\textbf{Technical Preview \& Contributions.} To overcome these limitations and fully unlock the potentials of neural networks in high-performance channel decoder design, in this paper we propose a novel \textit{doubly residual neural} decoder, namely \textbf{DRN} decoder, to provide strong decoding performance with low storage and computational costs. As revealed by its name, a key feature of DRN decoder is its built-in residual characteristics on both data processing and network structure, which jointly avoid the structured limitations of the existing neural channel decoders. In overall, we summarize the contributions and benefits of DRN decoder as follows:

\begin{itemize}
    \item Inspired by the historical success of ResNet \cite{he2016deep}, DRN decoder imposes both \textbf{residual input} and \textbf{residual learning} on the neural channel decoder architecture. Such structure-level reformulation ensures that DRN decoder can effectively and consistently learn strong error-correcting capability over various types of channel codes with different code lengths and code rates.
    
    \item Our experimental results show that, our proposed DRN decoder achieves significant decoding performance improvement. Compared with the state-of-the-art NBP decoder, DRN decoder enjoys 0.5$\sim$1.8 dB extra coding gain over different channel codes. Compared with HGN decoder, which has the strongest error-correcting capability among all the existing neural channel decoders, DRN decoder also achieves similar or even better decoding performance over different channel codes.
    
    \item DRN decoder also enjoys low-cost benefits on both model size and computational demand. Compared with NBP decoder, DRN decoder requires 23$\times$$\sim$100$\times$ fewer parameters and 3.2$\times$$\sim$4.3$\times$ fewer computational operations. Compared with HGN decoder, DRN decoder achieves the similar decoding performance with only using 373$\times$$\sim$2725$\times$ fewer parameters and 708$\times$$\sim$30054$\times$ fewer computational operations over different channel codes.

\end{itemize}

\textbf{Focus on Block Codes.} Channel codes can be roughly categorized to two types: \textit{block codes} and \textit{convolutional codes}. This paper focuses efficient neural channel decoder design for block codes, including LDPC, Polar and BCH codes. This is because block codes are the state-of-the-art channel codes due to their better error-correcting performance and more feasible decoder implementation than the convolutional codes. Currently most advanced communication (e,g, 5G) and storage systems (e,g, SSD) adopts block codes in the industrial standards and commercial products.

\section{Background and Related Work}

\subsection{Classical BP-based Channel Decoder}

\textbf{Channel Codes.} In general, for an ($n$, $k$) channel code with $n$-bit code length and $k$-bit information length, it can be defined by a binary \textit{generator matrix} $\mathbf{G}$ of size $k\times n$. Meanwhile, it is also associated with a binary \textit{parity check} matrix $\mathbf{H}$ of size $(n-k)\times n$, where $\mathbf{GH^T=0}$. 

In encoding phase, the original $k$-bit binary information vector $\mathbf{m}$ is encoded to an $n$-bit binary codeword $\mathbf{x=mG}$, where all the arithmetic operations are in binary domain. After $\mathbf{x}$ is transmitted over a noisy channel, at the receiver end the received codeword $\mathbf{r}$ is observed, and the goal of channel decoding is to recover $\mathbf{x}$ from $\mathbf{r}$.
\footnote{In practice the encoder usually adopts systematic encoding strategy \cite{lin1983djc}, so after decoding phase $\mathbf{m}$ can be directly obtained via fetching the first $k$ bits of the decoded $\mathbf{\hat{x}}$.}

\begin{figure}[tb]
\centering
\begin{subfigure}{\columnwidth}
\centering
\includegraphics[clip, trim=7cm 5cm 6cm 5cm, width=\columnwidth]{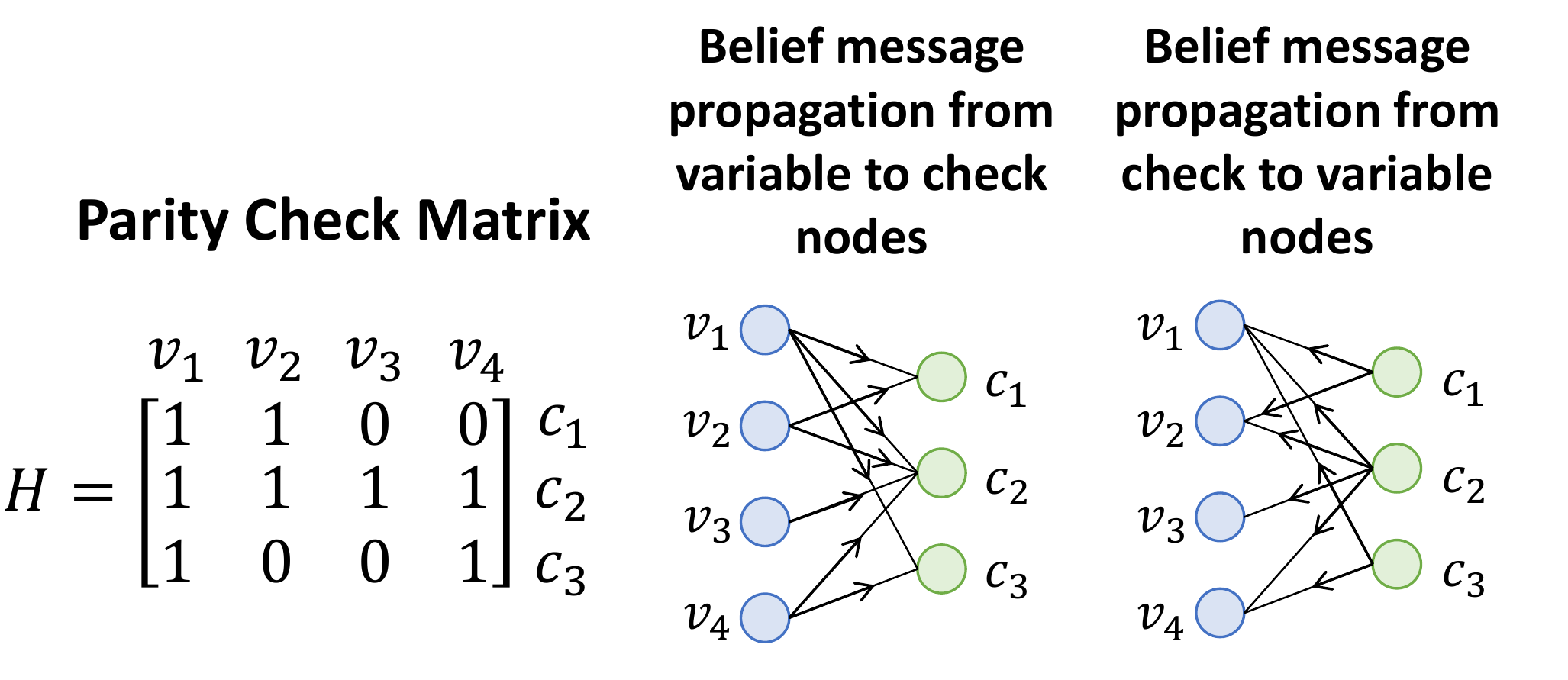}
\caption{}
\label{fig:tanner}
\end{subfigure}
\begin{subfigure}{.9\columnwidth}
\centering
\includegraphics[clip, trim=9cm 5cm 8.5cm 4.5cm, width=\columnwidth]{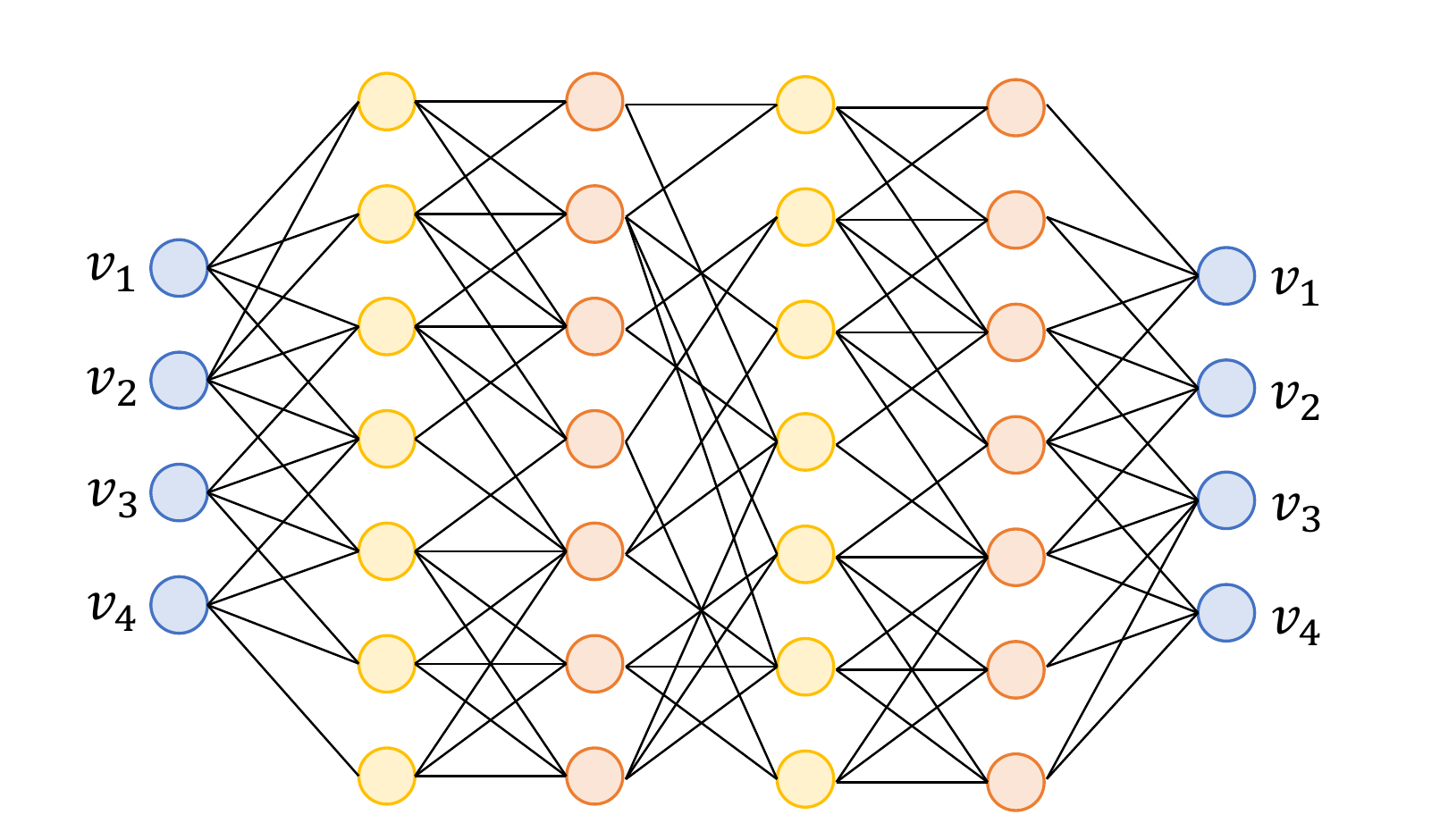}
\caption{}
\label{fig:trellis}
\end{subfigure}
\caption{(a) Parity check matrix and associate factor graph for channel codes. Iterative BP is on the factor graph. (b) Factor graph can be unfolded to Trellis graph. Variable nodes are colored in blue, V-to-C messages are in yellow, and C-to-V messages are in orange.}
\label{fig:tannerandtrellis}
\end{figure}

\textbf{Factor Graph and BP Algorithm.} Channel decoding can be performed by using various approaches. Among them, belief propagation (BP) is the most advanced decoding algorithm. The key idea of BP algorithm is to perform iterative belief message passing over the \textit{factor graph}, a bipartite graph entailed by parity check matrix $\mathbf{H}$. As illustrated in Figure \ref{fig:tanner}, the factor graph for an ($n$, $k$) channel code contains $n$ variable nodes and ($n-k$) check nodes, and each edge in the graph corresponds to an entry-1 in matrix $\mathbf{H}$. 

At the initial stage of BP algorithm, all the variable nodes receive the log likelihood ratio (LLR) $l_v$ of the corresponding bit:

\begin{equation}
l_v 
= 
\log\frac{P(x_v=1|\\r_v)}{P(x_v=0|\\r_v)},
\end{equation}
where $v\in[n]$ is the index of variable nodes, and $x_v$ and $r_v$ are the corresponding bit of $\mathbf{x}$ and $\mathbf{r}$, respectively. Then, the belief messages between variable nodes and check nodes are iteratively calculated and propagated as follows:

\begin{align}
\label{eq:bp}
\begin{split}
u_{v\rightarrow c}^t 
&= 
l_v + \sum_{c'\in N(v)\backslash c}u^{t-1}_{c'\rightarrow v}
\\
u_{c\rightarrow v}^t
&=
2\arctanh{\lbrack\prod_{v'\in M(c)\backslash v}\tanh{(\frac{u^t_{v'\rightarrow c} }{2})}\rbrack}
\\
s^t_v 
&= l_v + \sum_{c'\in N(v)}u_{c'\rightarrow v}^{t}
\end{split},
\end{align}
where $c\in[n-k]$ is the index of check nodes, and $t$ is the iteration number. $N(\cdot)$ and $M(\cdot)$ represent the set of the connected nodes to the current variable node and check node, respectively. $u_{c\rightarrow v}^t$ denotes the message to be propagated from the index-$c$ check node to the index-$v$ variable at the $t$-th iteration, and $u_{v\rightarrow c}^t$ denotes the message in the opposite direction. In addition, after the final iteration ($L$) $s^L_v$ is used for hard decision of the decoded bit $\hat{x}_v$. If $s^L_v > 0$, then $\hat{x}_v=1$; otherwise $\hat{x}_v=0$.

\subsection{Neural BP (NBP) Decoder}

From the perspective of neural network, the iterative BP decoding over factor graph can be "unfolded" to a neural network. Specifically, since the unfolded factor graph is essentially a Trellis graph, where each edge in the factor graph becomes the node of the Trellis graph (see Figure \ref{fig:trellis}), the entire Trellis graph can be interpreted as a special neural network, thereby forming a neural BP (NBP) decoder.  A very attractive advantage of this interpretation is that, with proper neural network training, each propagated message's associate scaling parameter, which was constant 1 or empirically set in conventional BP decoder, can now be trained as the weight of neural network to achieve better decoding performance. In general, the original message passing described in Eq. (\ref{eq:bp}) become the forward propagation on the layers of the NBP decoder \cite{nachmani2018deep} as follows:

\begin{align}
\label{eq:nbp}
\begin{split}
u_{v\rightarrow c}^t 
&= 
f(
w^t_{v,in} l_v + \sum_{c'\in N(v)\backslash c}
w^t_{c'\rightarrow v} u^{t-1}_{c'\rightarrow v})
\\
u_{c\rightarrow v}^t
&=
g(\prod_{v'\in M(c)\backslash v}u^t_{v'\rightarrow c})
\\
s^t_v 
&= \sigma(w^t_{v,out} l_v + \sum_{c'\in N(v)}
w^t_{c'\rightarrow v} u_{c'\rightarrow v}^{t})
\end{split},
\end{align}
where $f(\cdot)$, $g(\cdot)$ and $\sigma(\cdot)$ are the tanh, arctanh and sigmoid function, respectively. From the perspective of neural network, $w^t_{v,in}, w^t_{c'\rightarrow v}, w^t_{v,out}$ and $w^t_{c'\rightarrow v}$, can be learned by minimizing the multi-label binary classification loss as follows:
\begin{equation}
\label{eq:loss}
loss =
\sum_{v=1}^N
-\lbrack
x_v\log s_v + 
(1-x_v)\log(1-s_v)
\rbrack,
\end{equation}
where $s_v=s^L_v$ is the output of the last layer of NBP decoder. 

\subsection{Hyper Graph Neural (HGN) Decoder}
\label{subsec:hgn}
In \cite{nachmani2019hyper}, a hyper graph neural (HGN) decoder is proposed to further improve the performance of neural channel decoder. 
Beyond the weight-learning strategy adopted in the NBP decoder, at each iteration HGN decoder directly learns the belief message calculation and propagation schemes between check nodes and variable nodes. 
Specifically, the update of $u_{v\rightarrow c}^{t}$ is now learned and performed via a graph neural network as follows:
\begin{equation}
    u_{v\rightarrow c}^{t} = \text{GNN}(l_v, u_{c'\rightarrow v}^{t-1}), \forall c'\in N(v)\backslash c.
\end{equation}
Because it is found that training such graph neural network is quite challenging due to the large amount of possible updating schemes, HGN decoder further uses another neural network to learn and predict the weights for the graph neural network. 
In overall, unlike NBP decoder, HGN decoder adopts flexible belief message update scheme because of its "hyper-network" structure. Such flexibility is believed to bring significant decoding performance improvement over the fixed-scheme NBP decoder.

\begin{figure*}[tb]
\centering
\includegraphics[clip, trim=0cm 3cm 0cm 4cm, width=\textwidth]{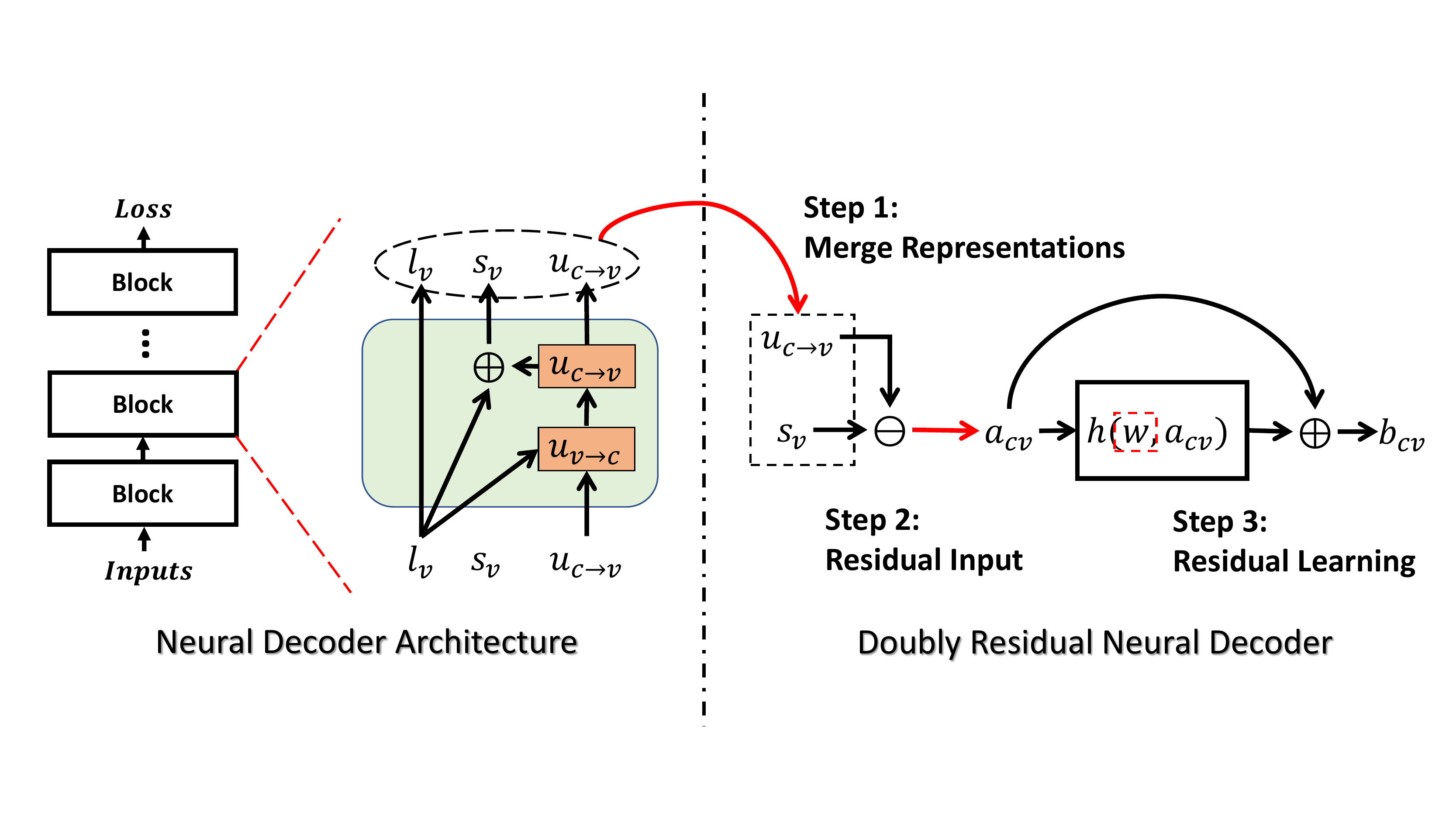}
\caption{Three-step reformulation to form DRN decoder.}
\label{fig:res}
\end{figure*}

\section{Method}

\subsection{Rethink and Analysis -- Lessons Learned from NBP and HGN Decoders}

\textbf{Dilemma between Performance and Cost.} Although NBP and HGN decoders show performance improvement over traditional BP decoder, they are facing several inherent limitations. For NBP decoder, its provided decoding performance improvement is not consistently significant. As will be shown in Section \ref{sec:exp}, on some channel codes (e.g. Polar codes) and with some codes parameters (e.g. higher code rate), the decoding performance of NBP decoder is similar to conventional BP decoder or even worse. On the other hand, HGN decoder shows consistently much lower BER with different types of codes and parameters. However, its unique hyper graph neural network structure makes it very expensive for both computation and storage. In overall, such dilemma between performance and cost severely hinders the widespread deployments of NBP and HGN decoders in practical applications.

\textbf{Rethink-1: Why is Performance of NBP Decoder Limited?} As mentioned above, the underlying design methodology used for NBP decoder -- training the unfolded factor graph as a neural network, though works, does not achieve the expected significant decoding performance improvement. We hypothesize such phenomenon is due to three reasons. 1) \underline{Depth.} Once factor graph is unfolded to Trellis graph, the depth of the corresponding neural network is proportional to the number of iterations, which is at least 5 in typically setting. Therefore, the depth of the NBP decoder is at least 10 layers or more. For such type of deep and plain neural network without additional structure such as residual block, it is well known that they suffer unsatisfied performance due to the vanishing gradient problem. 2) \underline{Sparsity.} Because factor graph of channel codes is inherently sparse, the underlying neural network of NBP decoder is highly sparse as well. Therefore, training an NBP decoder is essentially training a sparse neural network from scratch.
Unfortunately, extensive experiments in literature have shown that, the performance of a sparse model via training-from-scratch is usually inferior to the same-size one via pruning-from-dense \cite{li2016pruning,luo2017thinet,he2017channel,yu2018nisp}.
Such widely observed phenomenon probably also limits the performance of NBP decoder. 3) \underline{Application}. Different from most other applications, channel decoding has extremely strict requirement for accuracy. Its targeted bit error rate range is typically $10^{-3}$ and below. Therefore, even though learning the weights increases the classification accuracy, if such increase is not very significant, it will not translate to obvious decoding performance important in terms of BER or coding gain (dB). 

\textbf{Rethink-2: Is Flexible Message Update Scheme in HGN Decoder a Must?} As introduced in Section \ref{subsec:hgn}, HGN decoder uses high-complexity hyper graph neural network to directly learn the message update schemes instead of the weights only. In other words, both how the messages are calculated and propagated are now learnable and flexible. Although such flexibility is widely believed as the key enabler for the promising performance of HGN decoder, we argue its necessity for the high-performance neural channel decoder design. Recall the structure of the state-of-the-art convolutional neural networks (CNNs), such as ResNet \cite{he2016deep} and DenseNet \cite{huang2017densely}, we can find that the propagation path of the information during both inference and training phases are not flexible but always fixed. 
Although there are a set of works studying "adaptive inference" \cite{bolukbasi2017adaptive,wang2018skipnet,hu2019dynamic}, the main benefit of introducing such flexibility is to accelerate inference speed instead of improving accuracy-- actually those adaptive inference work typically have to trade the accuracy for faster inference. 

\textbf{Rethink-3: How to Break Performance-Cost Dilemma?} Based on our above analysis and observation, we believe designing a high-performance low-complexity neural channel decoder is not only possible, but the avenue is already available -- a new network architecture is the key. 
This is because the history of developing advanced CNNs, such as ResNet and DenseNet, has already demonstrated how important a new, instead of flexible, network architecture to the accuracy performance of CNN models. 
Inspired by these historical success, we propose to perform architecture-level reformulation to NBP decoder. Such design strategy is attractive for breaking the performance-cost dilemma of neural channel decoder because 1) NBP decoder itself has lower complexity than HGN decoder; and 2) if properly performed, architecture reformulation will bring high decoding performance. 

\subsection{Doubly Residual Neural (DRN) Decoder} 
\label{subsec:drn}

\textbf{Residual Structure: From CNN to Channel Decoder.} To achieve that, we propose to integrate \textit{residual structure}, which is a key enabler for the success of ResNet in CNN, to the design of high-performance neural channel decoder. 
As analyzed and verified by numerous prior studies, the residual structure, performs residual learning to learn the residual mapping  $\mathcal{F}(\mathbf{x}):=\mathcal{H}(\mathbf{x})-\mathbf{x}$ instead of directly learning the underlying mapping $\mathcal{H}(\mathbf{x})$. 
Such strategy effectively circumvents the vanishing gradient problem and makes training high-performance deep network become possible. As analyzed in our rethinking on the limitations of NBP decoder, such benefit provided by the residual structure is particular attractive for high-performance neural channel decoder design.

\textbf{Doubly Residual Structure.} Next we describe the proposed architecture reformulation on the neural channel decoder. As shown in Figure \ref{fig:res}, the entire decoder consists of multiple blocks, where each block stacks two adjacent layers of Trellis graph. Similar to the construction of bottleneck block in ResNet, our architecture reformulation is performed on this two-layer-stacked component block of the decoder.

\textbf{\underline{Mapping Challenge.}} Imposing the residual structure on the block is facing a structure-level challenge. For each component block, it maps three inputs to three outputs. From the perspective of DNN, such multiple-input-to-multiple-output mapping is very difficult for the neural network model to learn properly and accurately, which then would significantly limiting the learning capability.

\begin{figure}[bt]
\centering
\includegraphics[width=\columnwidth]{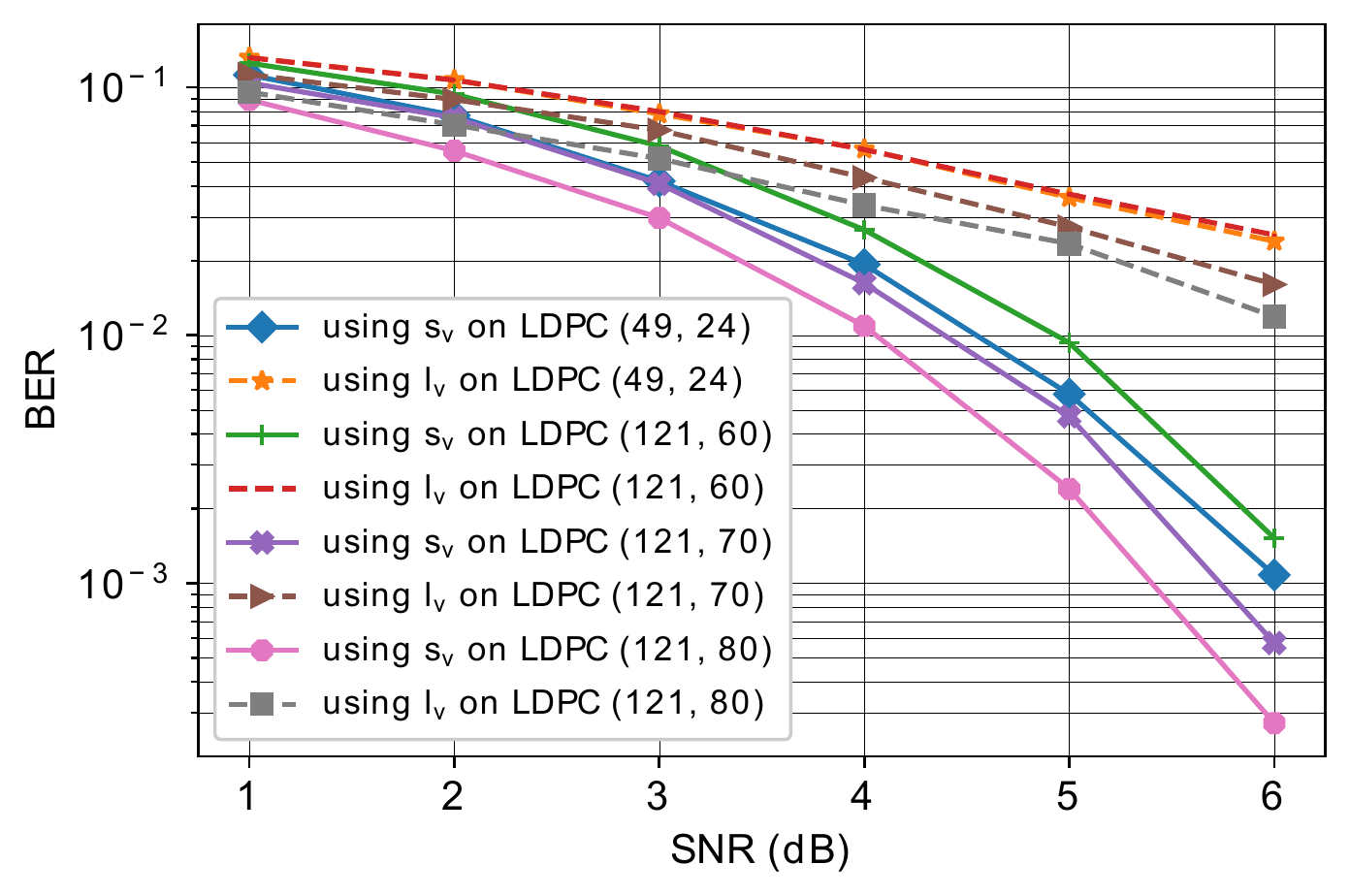}
\caption{BER of BP decoder using $s_v$ and $l_v$ for hard decision after 1 iteration on different LDPC codes.}
\label{fig:llr}
\end{figure}

\textbf{\underline{Step-1: Merge Representations.}} To overcome this challenge, we propose to simplify the input-to-output mapping in each block (see Figure \ref{fig:res}). Our first step is to merge $s_v$ and $l_v$ -- we use $s_v$ to replace $l_v$ in the corresponding computation. Such substitution is based on the phenomenon that, as the soft output for each iteration, $s_v$ should always be more reliable for hard decision of each bit than $l_v$, since $l_v$ is only the constant extrinsic LLR obtained from the noisy channel. For instance, as shown in Figure \ref{fig:llr}, when we simply use $l_v$ and $s_v$ after one BP iteration for hard decision of different LDPC codes, the BER performance using $s_v$ is much better than that using $l_v$ for hard decision. Based on this observation, in our proposed design we merge $s_v$ and $l_v$ at each block and only use $s_v$ for the involved computation.

\textbf{\underline{Step-2: Residual Input.}} After merging $s_v$ with $l_v$, there still exist 2-input-to-2-output mapping in the block. Hence we further propose to only use the residual value between $s_v$ and $u_{c\rightarrow v}$ as the input and output as follows:
\begin{equation}
a_{cv} = s_v - u_{c\rightarrow v}.
\end{equation}
As shown in Figure \ref{fig:res}, making residual input ensures that the component block only need to learn one-to-one mapping, thereby reducing the learning difficulty.

\textbf{\underline{Step-3: Residual Learning.}} Based on the one-to-one mapping result from the previous two steps, we can now integrate the shortcut-based residual learning to the decoder architecture. In general, the reformulated block will learn the following mapping function:
\begin{equation}
\begin{split}
b_{cv} 
&= a_{cv} + h(w, a_{cv'})
\\
&= a_{cv} + g\circ f(w, a_{cv'}),
\end{split}
\end{equation}
where $h(\cdot)$ is the activation function as the composition of $g(\cdot)$ and $f(\cdot)$. Figure \ref{fig:res} shows the overall procedure of this 3-step architecture reformulation. Since this new structure contains both residual input and residual learning, we name the entire decoder as doubly residual neural (DRN) decoder. 

\textbf{\underline{Further Complexity Reduction.}} 
Besides architecture reformulation, we also adopt two approaches to further reduce complexity of DRN decoder. 
First, during the training phase we keep the weights in the same block as the same. 
Our experimental results shows that, such weight sharing strategy significantly degrades the decoding performance of NBP decoder, but it does not affect DRN decoder at all. 
Second, considering the high complexity of tanh and arctanh functions in Eq. (\ref{eq:bp}), we adopt the widely used min-sum approximation \cite{hu2001efficient} to simplify the computation:
\begin{equation}
\begin{split}
y 
&= 2\arctanh[\tanh(\frac{p}{2})\tanh(\frac{q}{2})] 
\\
&\approx
\text{sign}(p)\cdot\text{sign}(q)\cdot\min(|p|,|q|),
\end{split}
\end{equation}
where $|\cdot|$ returns the absolute value. 
Based on this approximation, $h(\cdot)$ can be performed as follows:
\begin{align}
\label{eq:approx}
h(w, a_{cv})=
w \min_{v'\in M(c)\backslash v}|a_{cv'}|\prod_{v'\in M(c)\backslash v}\text{sign}{(a_{cv'})}.
\end{align}

\begin{table*}[tb]
\centering
\begin{tabular}{r|ccc|ccc|ccc|ccc}
\hline
Decoder & 
\multicolumn{3}{c|}{Conventional BP} & 
\multicolumn{3}{c|}{NBP} & 
\multicolumn{3}{c|}{HGN (NeurIPS'19)} & 
\multicolumn{3}{c}{DRN (Ours)} 
\\ \cline{1-13} 
SNR (dB) & 4 & 5 & 6 & 4 & 5 & 6 & 4 & 5 & 6 & 4 & 5 & 6 \\ \hline
Polar (64, 32) &
4.45 & 5.41 & 6.46 &
4.48 & 5.35 & 6.50 &
4.25 & 5.49 & 7.02 &
\bf{6.00} & \bf{7.97} & \bf{10.39}
\\
Polar (64, 48) &
4.64 & 5.90 & 7.31 &
4.52 & 5.73 & 7.49 &
4.91 & 6.48 & 8.41 &
\bf{5.80} & \bf{7.54} & \bf{10.03}
\\
Polar (128, 64) &
3.74 & 4.43 & 5.64 &
3.67 & 4.63 & 5.85 & 
3.89 & 5.18 & 6.94 &
\bf{5.32} & \bf{7.23} & \bf{9.67}
\\
Polar (128, 86) &
3.94 & 4.87 & 6.24 &
3.96 & 4.88 & 6.20 &
4.57 & 6.18 & 8.27 &
\bf{5.34} & \bf{6.92} & \bf{8.92}
\\
Polar (128, 96) &
4.13 & 5.21 & 6.43 &
4.25 & 5.09 & 6.75 &
4.73 & 6.39 & 8.57 &
\bf{5.40} & \bf{7.22} & \bf{9.60}
\\ \hline
LDPC (49, 24) &
5.36 & 7.26 & 10.03 &
5.29 & 7.67 & 10.27 &
5.76 & \bf{7.90} & 11.17 &
\bf{5.77} & 7.86 & \bf{11.28}
\\
LDPC (121, 60) &
4.76 & 7.20 & 11.07 &
4.96 & 8.00 & 12.35 &
5.22 & 8.29 & 13.00 &
\bf{5.26} & \bf{8.37} & \bf{13.20}
\\
LDPC (121, 70) &
5.85 & 8.93 & 13.75 &
6.43 & 9.53 & 13.83 &
6.39 & 9.81 & 14.04 &
\bf{6.39} & \bf{10.10} & \bf{15.43}
\\
LDPC (121, 80) &
6.54 & 9.64 & 14.78 &
7.04 & 10.56 & 14.97 &
6.95 & 10.68 & 15.80 &
\bf{7.31} & \bf{11.24} & \bf{17.00}
\\ \hline
BCH (31, 16) &
4.44 & 5.78 & 7.31 &
4.84 & 6.34 & 8.20 &
\bf{5.05} & \bf{6.64} & \bf{8.80} &
4.93 & 6.57 & 8.76
\\
BCH (63, 36) &
3.58 & 4.34 & 5.29 &
4.02 & 5.33 & 6.89 &
3.96 & \bf{5.35} & 7.20 &
\bf{4.10} & 5.33 & \bf{7.23}
\\
BCH (63, 45) &
3.84 & 4.92 & 6.35 &
4.37 & 5.61 & 7.20 &
4.48 & \bf{6.07} & \bf{8.45} &
\bf{4.53} & 5.97 & 8.16
\\ 
BCH (63, 51) &
4.21 & 5.32 & 6.75 & 
4.44 & 5.85 & 7.44 &
4.64 & 6.08 & 8.16 &
\bf{4.76} & \bf{6.21} & \bf{8.27}
\\ \hline
\end{tabular}
\caption{Negative logarithm of BER performance of different neural channel decoders. \textbf{High value means better performance.}}
\label{tbl:ber}
\end{table*}

\begin{table}[tb]
\centering
\begin{tabular}{r|c|c|c}
\hline
& NBP & HGN & DRN (Ours)
\\\hline
Polar (64, 32) & 41.1KB & 596.3KB & 1.6KB \\
Polar (64, 48) & 32.7KB & 428.0KB & 860B \\
Polar (128, 64) & 88.6KB & 1.4MB & 3.8KB \\
Polar (128, 86) & 111.5KB & 1.4MB & 3.3KB \\
Polar (128, 96) & 75.0KB & 1.0MB & 2.2KB \\ \hline
LDPC (49, 24) & 43.1KB & 447.6KB & 560B \\
LDPC (121, 60) & 246.8KB & 1.6MB & 1.3KB \\
LDPC (121, 70) & 193.6KB & 1.4MB & 1.1KB \\
LDPC (121, 80) & 145.2KB & 1.1MB & 880B \\ \hline
BCH (31, 16) & 30.9KB & 281.2KB & 300B \\
BCH (63, 36) & 269.4KB & 1.1MB & 540B \\
BCH (63, 45) & 277.4KB & 981.0KB & 360B \\
BCH (63, 51) & 229.4KB & 761.3KB & 240B \\
\hline
\end{tabular}
\caption{Model sizes of different neural channel decoders.}
\label{tbl:size}
\end{table}

\begin{table}[tb]
\centering
\begin{tabular}{r|c|c|c|c}
\hline
& BP & NBP & HGN & DRN\\\hline
Polar (64, 32) & 43.6K & 52.5K & 80.8M & 16.4K\\
Polar (64, 48) & 45.0K & 52.2K & 30.4M & 15.1K\\
Polar (128, 64) & 93.1K & 112.1K & 1.1G & 36.6K\\
Polar (128, 86) & 141.9K & 166.7K & 935.0M & 48.1K\\
Polar (128, 96) & 90.2K & 106.6K & 431.7M & 32.2K\\\hline
LDPC (49, 24) & 54.1K & 63.9K & 34.1M & 17.6K\\
LDPC (121, 60) & 316.4K & 374.5K & 1.6G & 94.4K\\
LDPC (121, 70) & 263.8K & 309.2K & 920.4M & 78.7K\\
LDPC (121, 80) & 211.1K & 245.0K & 476.1M & 62.9K\\\hline
BCH (31, 16) & 38.0K & 45.1K & 8.5M & 12.0K\\
BCH (63, 36) & 347.8K & 412.7K & 481.6M & 97.2K\\
BCH (63, 45) & 412.9K & 480.1K & 340.3M & 112.3K\\
BCH (63, 51) & 375.0K & 430.6K & 162.8M & 100.8K\\
\hline
\end{tabular}
\caption{FLOPs of different neural channel decoders to decode one codeword.}
\label{tbl:time}
\end{table}

\section{Experiment}
\label{sec:exp}
In this section, we compare DRN decoder with the traditional BP and the state-of-the-art NBP and HGN decoders in terms of decoding performance (BER), model size and computational cost. 

\subsection{Experimental Setting}

\textbf{Channel Codes Type.} All the decoders are evaluated on three types of popular ($n$,$k$) channel codes: LPDC, Polar and BCH codes with different code lengths and code rates. The parity check matrices are adopted from \cite{channelcodes}. 

\textbf{Iteration Number and Channel Condition.} For fairness the number of iterations for all the decoders is set as 5. Additive white Gaussian noise (AWGN) channel, as the mostly used channel type for channel coding research, is adopted for transmission channel. The signal-to-noise ratio (SNR) is set in the range of $1\sim6$dB.

\textbf{Experiment Environment.} Our experiment environment is Ubuntu 16.04 with 256GB random access memory (RAM), Intel(R) Xeon(R) CPU E5-2698 v4 @ 2.20GHz and Nvidia-V100 GPU.

\textbf{Training \& Testing.} Each input batch is mixed with equal number of samples from different SNR settings. 
The training batch size is 384, so there are 64 samples generated at each SNR value.  
We use the RMSprop optimizer \cite{hinton2012neural} with learning rate 0.001 and run 20,000 iterations. 
The training samples are generated on the fly and testing samples are generated till at least 100 error samples detected at each SNR setting.

\begin{figure*}[tb]
\begin{subfigure}[b]{0.33\textwidth}
    \centering
    \includegraphics[width=\textwidth]{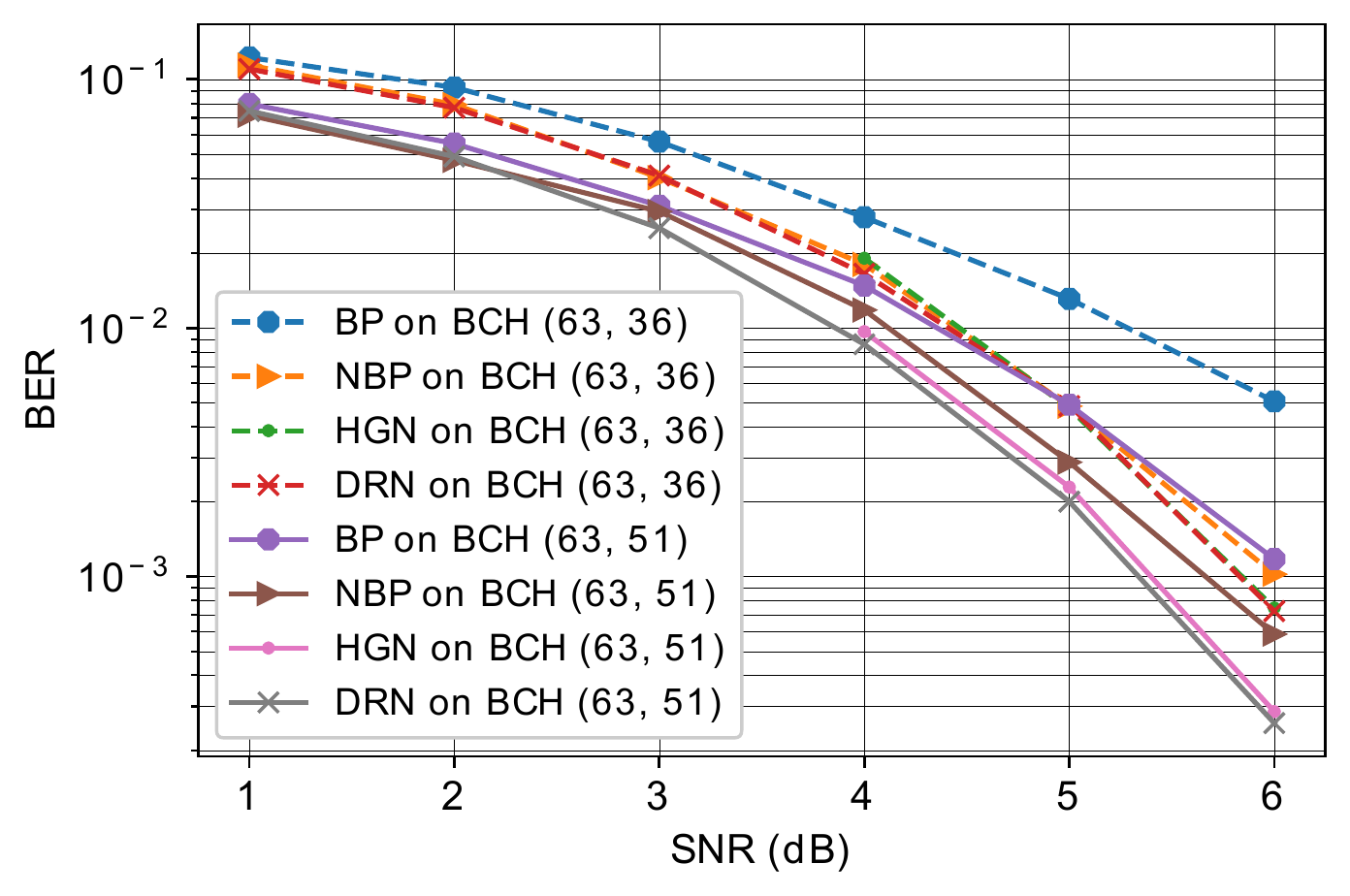}
    \caption{BCH codes with $n=63$.}    
    \label{fig:bch}
\end{subfigure}
\begin{subfigure}[b]{0.33\textwidth}  
    \centering 
    \includegraphics[width=\textwidth]{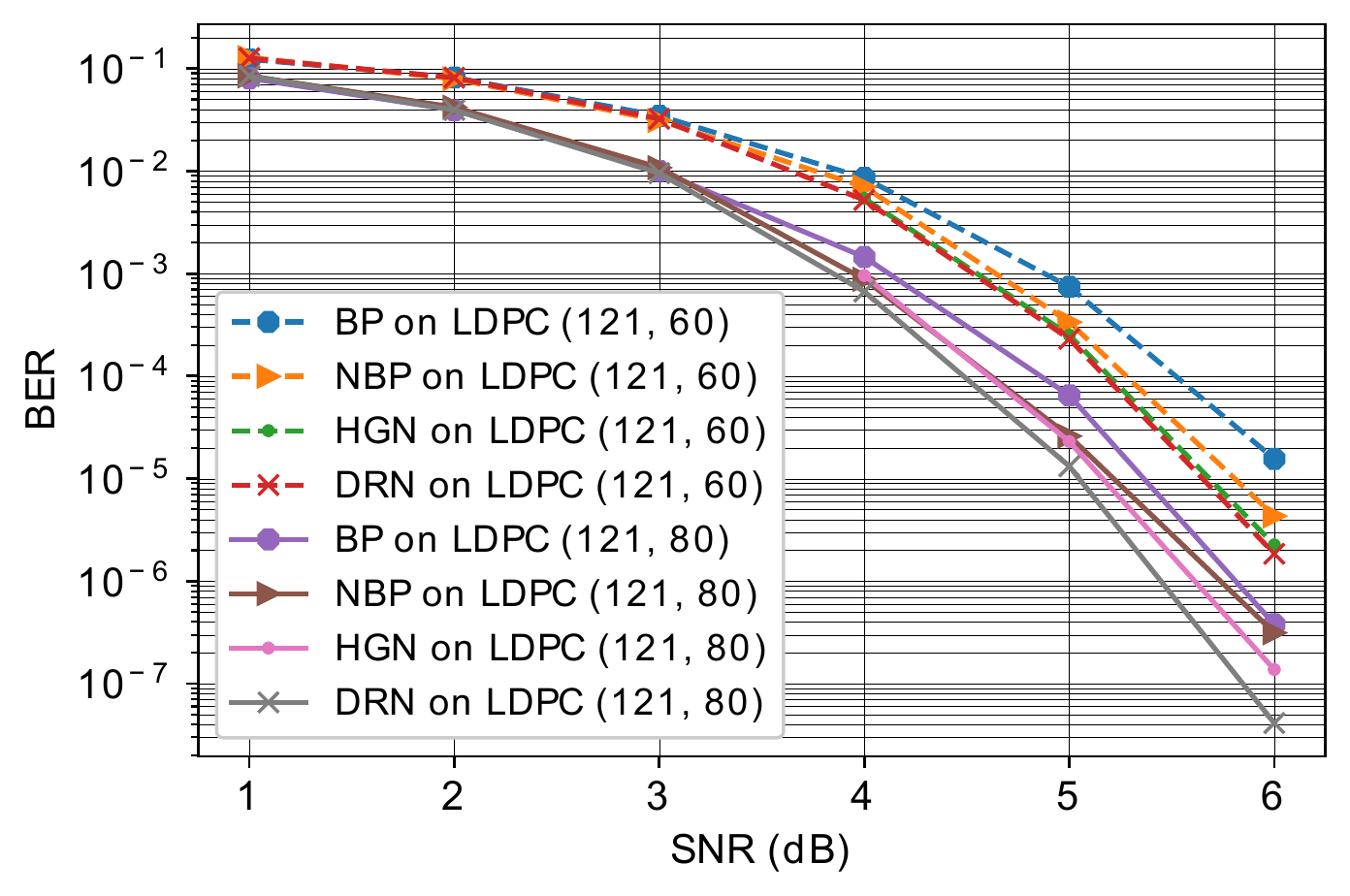}
    \caption{LDPC codes with $n=121$.}    
    \label{fig:ldpc}
\end{subfigure}
\begin{subfigure}[b]{0.33\textwidth}   
    \centering 
    \includegraphics[width=\textwidth]{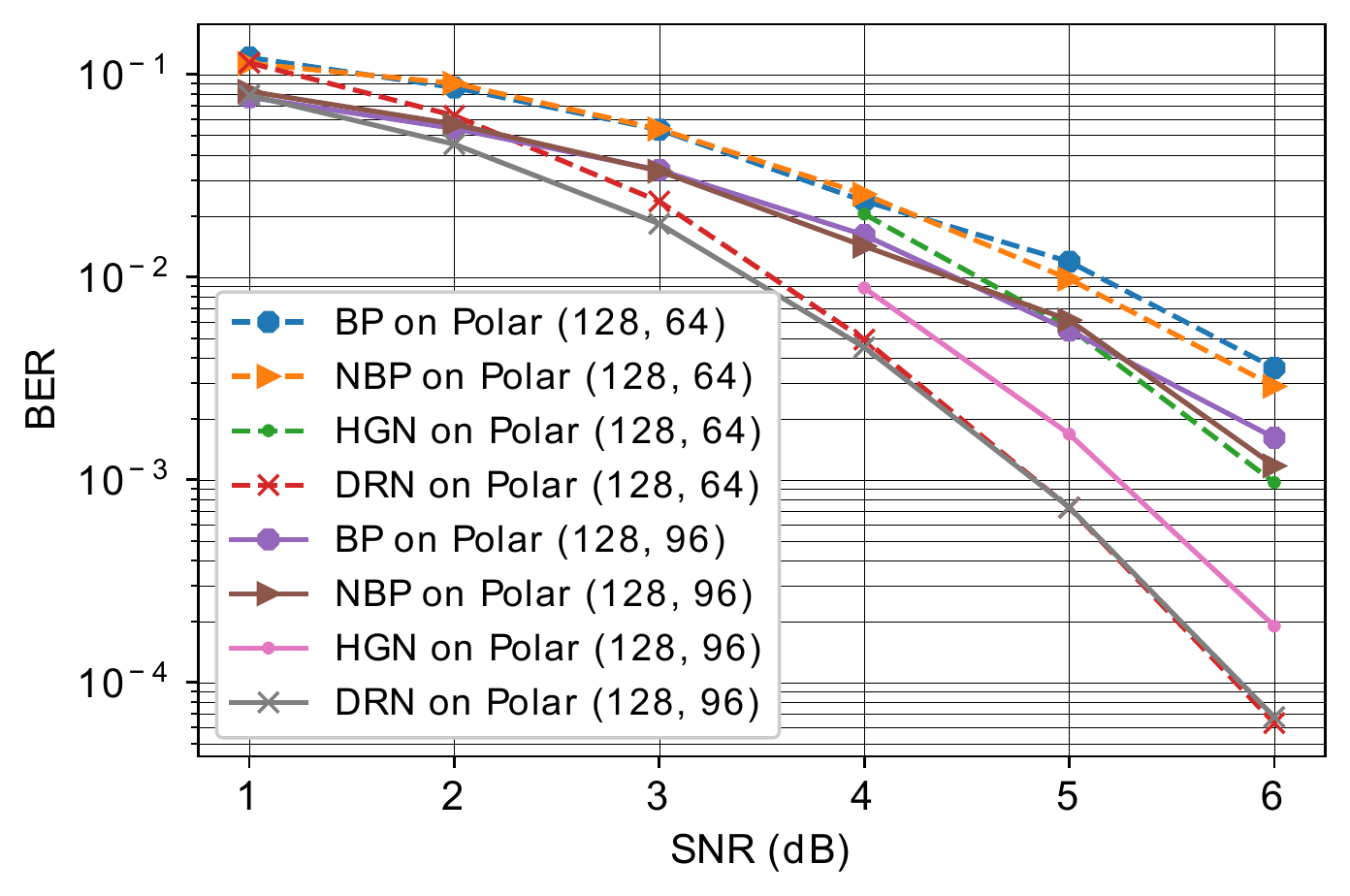}
    \caption{Polar codes with $n=128$.}    
    \label{fig:polar128}
\end{subfigure}
\caption
{BER-vs-SNR curve of different decoders on different channel codes.} 
\label{fig:ber}
\end{figure*}

\subsection{Decoding Performance (BER)}
Since BER can range from $10^{-1}$ to $10^{-8}$, for simplicity, we adopt the negative logarithm representation as used in HGN paper. Table \ref{tbl:ber} lists the negative logarithm of BER performance of different decoding methods. A higher number means a better performance because it corresponds to a lower BER. From this table it is seen that, with the built-in doubly residual structure, our DRN decoder obtain very strong error-correcting capability. It consistently achieves the best BER performance on most of Polar and LDPC codes. For BCH codes, DRG decoder achieve almost the same or better performance than HGN decoderl.

Figure \ref{fig:ber} shows the BER-vs-SNR curve for different decoders on different channel codes. Notice that HGN decoder only reports the BER under SNR=4$\sim$6dB. It can be seen that compared with the state-of-the-art NBP decoder, our DRN decoder consistently outperforms NBP decoder at all SNR settings (0.5$\sim$1.8 dB coding gain). Compared with the current most powerful HGN decoder, DRN decoder still achieves the similar or even better decoding performance over all the evaluated channel codes. 

Besides, compared with successive cancellation (SC) algorithm, which is a unique decoding approach for polar codes, DRN also shows better performance. 
For instance, on Polar (64, 32), SC has BER performance as 0.3 at 1dB, 0.14 at 2dB, 0.029 at 3dB, which are inferior to DRN. Though SC list (SCL) decoder can bring better BER performance, SCL suffers the inherent serial decoding scheme and linear increase in cost as list size increases.

\subsection{Model Size}
Table \ref{tbl:size} compares the model sizes of different decoders. Based on its inherent lightweight structure and weight sharing strategy, our DRN decoder requires the fewest model size than others over all different channel codes. 
Compared with NBP decoder, DRN decoder brings 23$\times$$\sim$100$\times$
reduction on model size.
Notice that as mentioned in Section \ref{subsec:drn}, the weight sharing strategy cannot be applied to NBP due to the resulting severe decoding performance loss. 
Also, compared with the large-size hyper graph neural network-based HGN decoder, DRN decoder enables 373$\times$$\sim$2725$\times$
reduction on model size with achieving the similar or better decoding performance as shown in Table \ref{tbl:ber} and Figure \ref{fig:ber}.

\subsection{Computational Cost}
Table \ref{tbl:time} compares the computational cost, in term of floating point operations (FLOPs) for decoding one codeword among different decoders. 
It can be seen that DRN decoder also enjoys the lowest computational cost because of its small-size model. 
Compared with NBP decoder, DRN decoder has 3.2$\times$$\sim$4.3$\times$
fewer computational cost. 
Compared with HGN decoder, DRN decoder needs  708$\times$$\sim$30054$\times$ fewer operations while achieving the same or better decoding performance.

\subsection{Analysis}
\begin{figure}[bt]
\centering
\includegraphics[width=\columnwidth]{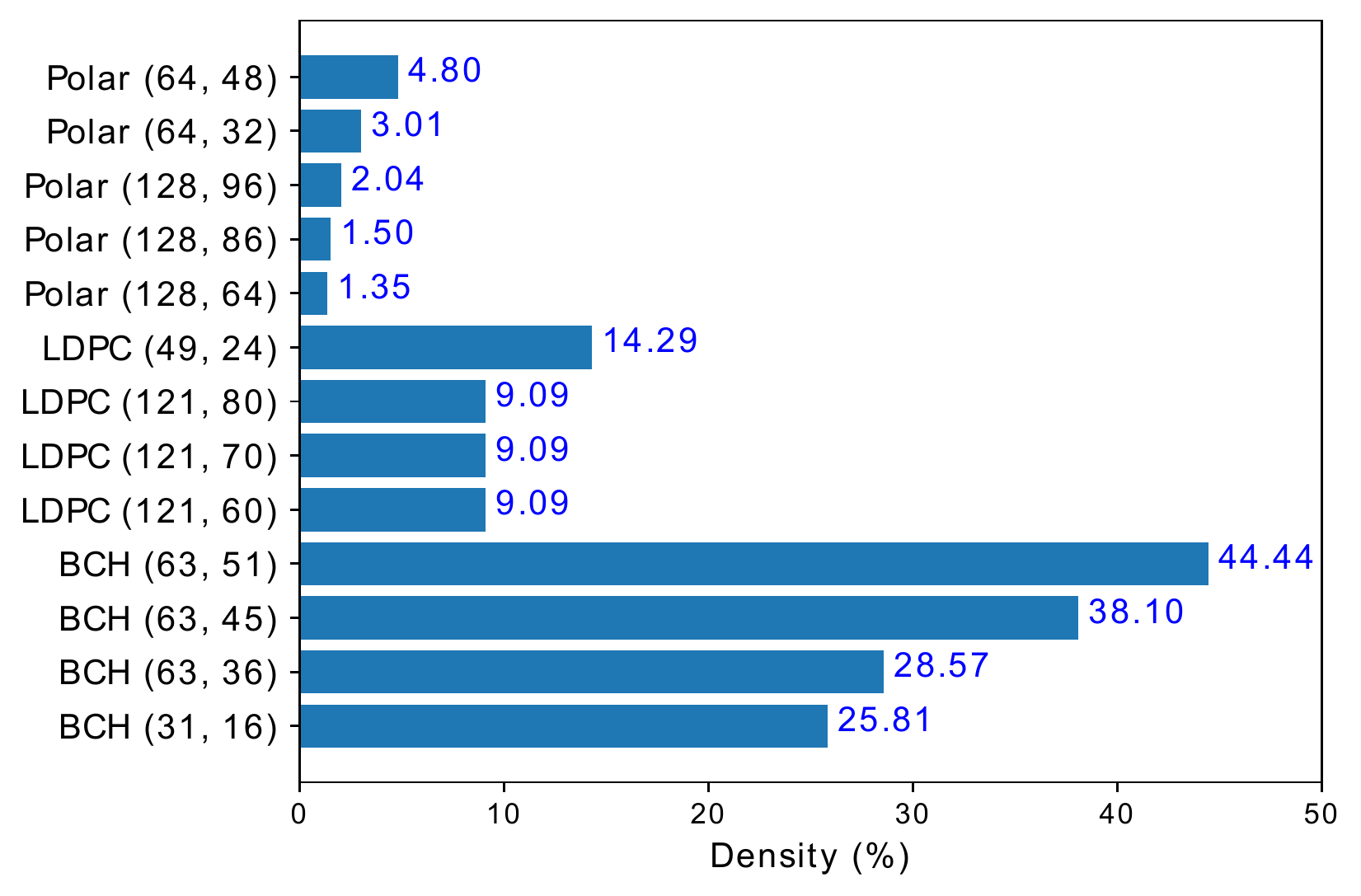}
\caption{Density of H matrix on different codes.}
\label{fig:dense}
\end{figure}

From simulation results it is seen that DRN achieves better BERs than HGN on 6 BCH codes, and achieves very close BERs on other 6 BCH codes. Though such performance is already very promising, the performance improvement over HGN is not as huge as that on LDPC and polar codes. We hypothesize such phenomenon is related to the density of H matrix.
For BP-family decoders, like our DRN, H matrix density highly affects BER performance. 
Figure \ref{fig:dense} shows H matrices of the evaluated BCH codes have higher density than those of LDPC and Polar codes, hence this may explain why DRN performs better on LDPC and Polar codes than on BCH codes.

\section{Conclusion}
This paper proposes doubly residual neural (DRN) decoder, a low-complexity high-performance neural channel decoder. Built upon the inherent residual input and residual learning structure, DRN decoder achieves strong decoding performance with low storage cost and computational cost. Our evaluation on different channel codes shows that the proposed DRN decoder consistently outperforms the state-of-the-art neural channel decoders in terms of decoding performance, model size and computational cost. 

\section{Acknowledgement}
This work is partially supported by National Science Foundation (NSF) award CCF-1854737.

\bibliography{ref}

\begin{thebibliography}{31}
\providecommand{\natexlab}[1]{#1}
\providecommand{\url}[1]{\texttt{#1}}
\providecommand{\urlprefix}{URL }
\expandafter\ifx\csname urlstyle\endcsname\relax
  \providecommand{\doi}[1]{doi:\discretionary{}{}{}#1}\else
  \providecommand{\doi}{doi:\discretionary{}{}{}\begingroup
  \urlstyle{rm}\Url}\fi

\bibitem[{Arikan(2009)}]{arikan2009channel}
Arikan, E. 2009.
\newblock Channel polarization: A method for constructing capacity-achieving
  codes for symmetric binary-input memoryless channels.
\newblock \emph{IEEE Transactions on information Theory} 55(7): 3051--3073.

\bibitem[{Bolukbasi et~al.(2017)Bolukbasi, Wang, Dekel, and
  Saligrama}]{bolukbasi2017adaptive}
Bolukbasi, T.; Wang, J.; Dekel, O.; and Saligrama, V. 2017.
\newblock Adaptive neural networks for efficient inference.
\newblock In \emph{Proceedings of the 34th International Conference on Machine
  Learning-Volume 70}, 527--536.

\bibitem[{Burth~Kurka and G{\"u}nd{\"u}z(2020)}]{burth2020joint}
Burth~Kurka, D.; and G{\"u}nd{\"u}z, D. 2020.
\newblock Joint source-channel coding of images with (not very) deep learning.
\newblock In \emph{International Zurich Seminar on Information and
  Communication (IZS 2020). Proceedings}, 90--94. ETH Zurich.

\bibitem[{Cammerer et~al.(2017)Cammerer, Gruber, Hoydis, and
  Ten~Brink}]{cammerer2017scaling}
Cammerer, S.; Gruber, T.; Hoydis, J.; and Ten~Brink, S. 2017.
\newblock Scaling deep learning-based decoding of polar codes via partitioning.
\newblock In \emph{GLOBECOM 2017-2017 IEEE Global Communications Conference},
  1--6. IEEE.

\bibitem[{Ebada et~al.(2019)Ebada, Cammerer, Elkelesh, and ten
  Brink}]{ebada2019deep}
Ebada, M.; Cammerer, S.; Elkelesh, A.; and ten Brink, S. 2019.
\newblock Deep learning-based polar code design.
\newblock In \emph{2019 57th Annual Allerton Conference on Communication,
  Control, and Computing (Allerton)}, 177--183. IEEE.

\bibitem[{Fossorier, Mihaljevic, and Imai(1999)}]{fossorier1999reduced}
Fossorier, M.~P.; Mihaljevic, M.; and Imai, H. 1999.
\newblock Reduced complexity iterative decoding of low-density parity check
  codes based on belief propagation.
\newblock \emph{IEEE Transactions on communications} 47(5): 673--680.

\bibitem[{Gallager(1962)}]{gallager1962low}
Gallager, R. 1962.
\newblock Low-density parity-check codes.
\newblock \emph{IRE Transactions on information theory} 8(1): 21--28.

\bibitem[{Gruber et~al.(2017)Gruber, Cammerer, Hoydis, and ten
  Brink}]{gruber2017deep}
Gruber, T.; Cammerer, S.; Hoydis, J.; and ten Brink, S. 2017.
\newblock On deep learning-based channel decoding.
\newblock In \emph{2017 51st Annual Conference on Information Sciences and
  Systems (CISS)}, 1--6. IEEE.

\bibitem[{He et~al.(2016)He, Zhang, Ren, and Sun}]{he2016deep}
He, K.; Zhang, X.; Ren, S.; and Sun, J. 2016.
\newblock Deep residual learning for image recognition.
\newblock In \emph{Proceedings of the IEEE conference on computer vision and
  pattern recognition}, 770--778.

\bibitem[{He, Zhang, and Sun(2017)}]{he2017channel}
He, Y.; Zhang, X.; and Sun, J. 2017.
\newblock Channel pruning for accelerating very deep neural networks.
\newblock In \emph{Proceedings of the IEEE International Conference on Computer
  Vision}, 1389--1397.

\bibitem[{Helmling et~al.(2019)Helmling, Scholl, Gensheimer, Dietz, Kraft,
  Ruzika, and Wehn}]{channelcodes}
Helmling, M.; Scholl, S.; Gensheimer, F.; Dietz, T.; Kraft, K.; Ruzika, S.; and
  Wehn, N. 2019.
\newblock {D}atabase of {C}hannel {C}odes and {ML} {S}imulation {R}esults.
\newblock \url{www.uni-kl.de/channel-codes}.

\bibitem[{Hershey, Roux, and Weninger(2014)}]{hershey2014deep}
Hershey, J.~R.; Roux, J.~L.; and Weninger, F. 2014.
\newblock Deep unfolding: Model-based inspiration of novel deep architectures.
\newblock \emph{arXiv preprint arXiv:1409.2574} .

\bibitem[{Hinton, Srivastava, and Swersky(2012)}]{hinton2012neural}
Hinton, G.; Srivastava, N.; and Swersky, K. 2012.
\newblock Neural networks for machine learning lecture 6a overview of
  mini-batch gradient descent .

\bibitem[{Hu et~al.(2019)Hu, Bao, Wang, and Liu}]{hu2019dynamic}
Hu, C.; Bao, W.; Wang, D.; and Liu, F. 2019.
\newblock Dynamic adaptive DNN surgery for inference acceleration on the edge.
\newblock In \emph{IEEE INFOCOM 2019-IEEE Conference on Computer
  Communications}, 1423--1431. IEEE.

\bibitem[{Hu et~al.(2001)Hu, Eleftheriou, Arnold, and
  Dholakia}]{hu2001efficient}
Hu, X.-Y.; Eleftheriou, E.; Arnold, D.-M.; and Dholakia, A. 2001.
\newblock Efficient implementations of the sum-product algorithm for decoding
  LDPC codes.
\newblock In \emph{GLOBECOM'01. IEEE Global Telecommunications Conference (Cat.
  No. 01CH37270)}, volume~2, 1036--1036E. IEEE.

\bibitem[{Huang et~al.(2017)Huang, Liu, Van Der~Maaten, and
  Weinberger}]{huang2017densely}
Huang, G.; Liu, Z.; Van Der~Maaten, L.; and Weinberger, K.~Q. 2017.
\newblock Densely connected convolutional networks.
\newblock In \emph{Proceedings of the IEEE conference on computer vision and
  pattern recognition}, 4700--4708.

\bibitem[{Jiang et~al.(2019)Jiang, Kim, Asnani, Kannan, Oh, and
  Viswanath}]{jiang2019turbo}
Jiang, Y.; Kim, H.; Asnani, H.; Kannan, S.; Oh, S.; and Viswanath, P. 2019.
\newblock Turbo autoencoder: Deep learning based channel codes for
  point-to-point communication channels.
\newblock In \emph{Advances in Neural Information Processing Systems},
  2758--2768.

\bibitem[{Kim et~al.(2018)Kim, Jiang, Kannan, Oh, and
  Viswanath}]{kim2018deepcode}
Kim, H.; Jiang, Y.; Kannan, S.; Oh, S.; and Viswanath, P. 2018.
\newblock Deepcode: Feedback codes via deep learning.
\newblock In \emph{Advances in neural information processing systems},
  9436--9446.

\bibitem[{Kim, Oh, and Viswanath(2020)}]{kim2020physical}
Kim, H.; Oh, S.; and Viswanath, P. 2020.
\newblock Physical Layer Communication via Deep Learning.
\newblock \emph{IEEE Journal on Selected Areas in Information Theory} .

\bibitem[{Li et~al.(2016)Li, Kadav, Durdanovic, Samet, and
  Graf}]{li2016pruning}
Li, H.; Kadav, A.; Durdanovic, I.; Samet, H.; and Graf, H.~P. 2016.
\newblock Pruning filters for efficient convnets.
\newblock \emph{arXiv preprint arXiv:1608.08710} .

\bibitem[{Lin and Costello(1983)}]{lin1983djc}
Lin, S.; and Costello, D.~J. 1983.
\newblock \emph{Error Control Coding: Fundamentals and Applications}.
\newblock prentice Hall.

\bibitem[{Lugosch and Gross(2017)}]{lugosch2017neural}
Lugosch, L.; and Gross, W.~J. 2017.
\newblock Neural offset min-sum decoding.
\newblock In \emph{2017 IEEE International Symposium on Information Theory
  (ISIT)}, 1361--1365. IEEE.

\bibitem[{Luo, Wu, and Lin(2017)}]{luo2017thinet}
Luo, J.-H.; Wu, J.; and Lin, W. 2017.
\newblock Thinet: A filter level pruning method for deep neural network
  compression.
\newblock In \emph{Proceedings of the IEEE international conference on computer
  vision}, 5058--5066.

\bibitem[{MacKay and Neal(1996)}]{mackay1996near}
MacKay, D.~J.; and Neal, R.~M. 1996.
\newblock Near Shannon limit performance of low density parity check codes.
\newblock \emph{Electronics letters} 32(18): 1645--1646.

\bibitem[{Nachmani, Be'ery, and Burshtein(2016)}]{nachmani2016learning}
Nachmani, E.; Be'ery, Y.; and Burshtein, D. 2016.
\newblock Learning to decode linear codes using deep learning.
\newblock In \emph{2016 54th Annual Allerton Conference on Communication,
  Control, and Computing (Allerton)}, 341--346. IEEE.

\bibitem[{Nachmani et~al.(2018)Nachmani, Marciano, Lugosch, Gross, Burshtein,
  and Be’ery}]{nachmani2018deep}
Nachmani, E.; Marciano, E.; Lugosch, L.; Gross, W.~J.; Burshtein, D.; and
  Be’ery, Y. 2018.
\newblock Deep learning methods for improved decoding of linear codes.
\newblock \emph{IEEE Journal of Selected Topics in Signal Processing} 12(1):
  119--131.

\bibitem[{Nachmani and Wolf(2019)}]{nachmani2019hyper}
Nachmani, E.; and Wolf, L. 2019.
\newblock Hyper-graph-network decoders for block codes.
\newblock In \emph{Advances in Neural Information Processing Systems},
  2329--2339.

\bibitem[{O’Shea and Hoydis(2017)}]{o2017introduction}
O’Shea, T.; and Hoydis, J. 2017.
\newblock An introduction to deep learning for the physical layer.
\newblock \emph{IEEE Transactions on Cognitive Communications and Networking}
  3(4): 563--575.

\bibitem[{Shannon(1948)}]{shannon1948mathematical}
Shannon, C.~E. 1948.
\newblock A mathematical theory of communication.
\newblock \emph{The Bell system technical journal} 27(3): 379--423.

\bibitem[{Wang et~al.(2018)Wang, Yu, Dou, Darrell, and
  Gonzalez}]{wang2018skipnet}
Wang, X.; Yu, F.; Dou, Z.-Y.; Darrell, T.; and Gonzalez, J.~E. 2018.
\newblock Skipnet: Learning dynamic routing in convolutional networks.
\newblock In \emph{Proceedings of the European Conference on Computer Vision
  (ECCV)}, 409--424.

\bibitem[{Yu et~al.(2018)Yu, Li, Chen, Lai, Morariu, Han, Gao, Lin, and
  Davis}]{yu2018nisp}
Yu, R.; Li, A.; Chen, C.-F.; Lai, J.-H.; Morariu, V.~I.; Han, X.; Gao, M.; Lin,
  C.-Y.; and Davis, L.~S. 2018.
\newblock Nisp: Pruning networks using neuron importance score propagation.
\newblock In \emph{Proceedings of the IEEE Conference on Computer Vision and
  Pattern Recognition}, 9194--9203.

\end{thebibliography}
\end{document}